\documentclass[letterpaper]{article} 
\usepackage[preprint]{aaai2027}
\usepackage{times}  
\usepackage{helvet}  
\usepackage{courier}  
\usepackage[hyphens]{url}  
\usepackage{graphicx} 
\usepackage{natbib}  
\usepackage{caption} 
\DeclareCaptionStyle{ruled}{labelfont=normalfont,labelsep=colon,strut=off}
\usepackage{booktabs}
\usepackage{amsmath}
\usepackage{amssymb}

\title{Why2Speak: Faithful Reasoning for Abstaining Action Policies}

\author{Shreya Mendi, Brinnae Bent}
\affiliations{Pratt School of Engineering, Duke University \\}

\begin{document}
\maketitle

\begin{abstract}
Many agentic systems must repeatedly choose between acting and abstaining, making faithful reasoning important for oversight: an explanation is useful only if it reflects the computation that produced the action, including the decision to do nothing. We study this problem through intervention timing in multi-party conversation, a controlled action-policy setting in which an assistant must decide whether to speak now or remain silent. This setting exposes challenges that are largely absent from question-answering faithfulness benchmarks, including severe class imbalance, asymmetric action costs, and the possibility that exposing reasoning changes the policy being audited. Using Qwen3-8B as a hybrid model that can be decoded with or without chain-of-thought reasoning, we compare direct decision policies, reasoning policies, supervised fine-tuning, and reinforcement learning. We find a consistent capability–auditability tradeoff: the strongest direct policy achieves higher decision quality but exposes no reasoning to inspect, while the reasoning policy provides an auditable trace at the cost of lower performance, particularly in recall of true intervention opportunities. Attempts to close this gap with supervised fine-tuning either suppress reasoning or preserve reasoning without improving decision quality. Reinforcement learning also fails to improve the reasoning policy; we identify a mechanism underlying this failure: group-relative objectives provide no learning signal on confidently wrong prompts when sampled rollouts all select the same action. We further audit the reasoning policy using controlled activation probes and behavioral ablations and show that standard faithfulness methods can overstate evidence that exposed reasoning reflects the underlying decision process. We show that standard faithfulness evaluations can be misleading in abstaining action policies: probability-based metrics saturate under confident decisions, probe-based analyses are vulnerable to class imbalance and textual leakage, and reasoning ablations can confound reasoning content with changes in inference mode. Together, these results show that exposing reasoning can change an agent’s action policy rather than simply make it observable. We characterize this capability–auditability tradeoff and provide methodological controls for evaluating reasoning-based oversight of agents that can act or abstain.

\end{abstract}

\section{Introduction}

AI agents are increasingly deployed in settings where action is optional. A
clinical assistant may flag a concern or remain silent; a tutoring system may
interrupt a student or allow productive struggle; a meeting assistant may add
missing information or let the conversation continue. In these domains, in
addition to producing a response, the agent must also decide whether
acting at all is appropriate. Unnecessary action can disrupt tasks and
erode trust, while inappropriate abstention can leave errors or missed
opportunities unaddressed.

This paper studies faithful reasoning for such abstaining action policies. We
focus on intervention timing in multi-party conversation as a concrete testbed:
given a transcript prefix, an assistant must decide whether to intervene now or
remain silent. Concurrent work under review shows that a learned classifier can identify when to
make a useful intervention significantly better than zero-shot prompting
\citep{when2speak}. However, that policy is opaque: it emits an action token but
provides no account of why that moment warrants speaking, or why silence is
preferable.

One solution is to require the policy to reason before acting. A conversational
agent could produce a short explanation such as ``the group has accepted an
incorrect deadline, so the agent should correct it,'' and then decide whether to
intervene. This would make the policy more inspectable to users and developers.
However, the explanation is valuable only if it is faithful: it must reflect
computation that actually influenced the action, rather than a plausible
rationalization generated after the decision was already made
\citep{jacovi2020}. Unfaithful explanations are especially risky in agentic
settings because they can make a system appear more transparent precisely when it
is not \citep{korbak2025,baker2025}.

Existing faithfulness work has primarily studied question answering, where models
always produce an answer and the answer can often be checked directly
\citep{turpin2023,lanham2023}. Abstaining action policies differ in three
important ways: first, the output is an action under asymmetric costs, not a
factual answer. Second, doing nothing is a meaningful outcome rather than missing
output. Third, the policy can be trained with reinforcement learning, so the
reward objective itself may affect whether reasoning becomes more or less
causally involved in the final action. These properties make intervention timing
a useful test case for studying faithful reasoning beyond QA-style answer
production.

We build a controlled testbed around a hybrid language model, Qwen3-8B
\citep{qwen3}, whose weights can be decoded either with explicit reasoning
(``think'' mode) or as a direct decision policy (``no-think'' mode). This design
allows us to compare reasoning and non-reasoning behavior without changing the
underlying model family or architecture. We then evaluate supervised fine-tuning,
reinforcement learning for accuracy, and reinforcement learning with a behavioral
faithfulness reward. Finally, we audit the resulting policies using activation
probes and behavioral interventions that remove or replace parts of the generated
reasoning.

This paper makes four main contributions:

\begin{enumerate}
\item We formulate faithfulness for an abstaining action policy. Intervention
timing requires the policy to justify not only what it says, but whether it
should act at all. This setting makes abstention, class imbalance, asymmetric
costs, and calibration first-order parts of the faithfulness problem.

\item We identify a capability--auditability tradeoff. The strongest decision
policy is a direct classifier that does not produce reasoning. The policy that
reasons is auditable but less accurate, with the largest cost appearing in missed
interventions.

\item We show that straightforward training does not remove the tradeoff.
Decision-token fine-tuning creates a stronger direct classifier but suppresses
reasoning. A masked objective that preserves the model's own reasoning cannot
improve beyond the base reasoning policy because it trains only on cases the
model already solves. RL in think mode also fails to substantially exceed the
base reasoning policy because group-relative objectives receive no learning
signal on prompts where all sampled rollouts agree, including confidently wrong
prompts.

\item We provide controlled audits and methodological guidance for action-policy
faithfulness. Under controlled probing, the reasoning policy's decision becomes
decodable during the chain of thought, while the direct classifier is already
decodable at the prompt. We also show that probability-based contribution
metrics, imbalance-blind probe accuracy, end-of-reasoning probes, and
uncontrolled flip tests can mislead in abstaining action-policy settings. We
further show that comparing reward objectives from a single training run per arm
can mistake seed variance for an objective effect.
\end{enumerate}

\section{Related Work}

\subsection{Intervention Timing in Multi-Party Dialogue}
Turn-taking has long been central to dialogue systems and human-robot interaction
\citep{skantze2021}. Intervention timing is a related but more specific problem:
an assistant observes an ongoing multi-party conversation and must decide whether
an external contribution would help at the current point. We build on a dataset for temporal participation and turn-taking introduced in concurrent work
\citep{when2speak}, in which synthetic conversations are constructed with epistemic gaps,
including factual correction, concept definition, data provision, source
identification, and synthesis or reframing. The task is to classify each
token-level decision point as intervene or wait.

\subsection{From QA Faithfulness to Action-Policy Faithfulness}
Most empirical work on chain-of-thought faithfulness studies answer production: a
model receives a question, generates reasoning, and then returns an answer.
Chain-of-thought prompting can improve performance and produce explanations that
appear useful to humans. However, prior work shows that generated reasoning is
often not faithful to the model's actual decision process. Models may follow
biased cues without mentioning them \citep{turpin2023}, reach the same answer
even when their reasoning is truncated or corrupted
\citep{lanham2023,chen2025}, or encode the final answer before the chain of
thought begins \citep{decodingbefore}.
Activation probes can reveal hidden signals that are absent from text, but probe
results require careful controls because decodability is not the same as causal
use \citep{hewitt2019,burns2022,hiddenerror}.

Our setting differs from standard question answering because the policy may
correctly remain silent. A faithfulness audit must therefore account for action
abstention, class imbalance, asymmetric error costs, and the possibility that
reasoning changes the policy's operating mode rather than the content of a
specific decision. These differences motivate treating intervention timing as an
action-policy faithfulness problem rather than as a direct application of
QA-style chain-of-thought auditing.

\section{Methods}

\subsection{Task and Data}
We evaluate faithfulness in an abstaining action-policy setting using the
benchmark developed in concurrent work \citep{when2speak}. The task is binary intervention timing:
given a prefix of a multi-party conversation, the model predicts whether an
assistant should intervene at the current point or remain silent.

The dataset contains approximately 173{,}000 token-level decision points derived
from approximately 16{,}000 synthetic conversations. Intervention opportunities
are rare, comprising roughly 13\% of examples, making the task highly
class-imbalanced. Each source scenario belongs to one of five intervention
types: factual correction, concept definition, data provision, source
identification, or synthesis and reframing. Although intervention type is not
included in the released token-level dataset, each conversation inherits an
identifier from its originating source scenario. We recover intervention types by
mapping conversations back to their source items. These recovered labels are used
only for interpretability analyses and never for supervising intervention
prediction.

During evaluation, the model predicts one of two actions: intervene or remain
silent. In no-think mode, the model emits only the decision token. In think mode,
it first generates a chain of thought and then outputs the final decision. Unless
otherwise stated, evaluation uses only the final decision token. Implementation
details, prompt templates, and dataset statistics are provided in Appendix~A.

\subsection{Evaluation Metrics}
Our primary deployment metric is macro-F1, which weights intervention and
non-intervention performance equally despite the severe class imbalance. We
additionally report two deployment-oriented error rates: the
\emph{false-intervention rate}, the fraction of silent moments on which the
assistant intervenes, and the \emph{missed-intervention rate}, the fraction of
true intervention opportunities on which the assistant remains silent
\citep{elkan2001}.

Because several training procedures change the intervention threshold, we also
report AUROC as a threshold-free diagnostic. Throughout the paper, AUROC is used
only to distinguish changes in discrimination from changes in calibration;
deployment performance is always evaluated using each policy's natural operating
point rather than \emph{post hoc} threshold tuning.

\subsection{Models and Training Conditions}
Our primary experiments use Qwen3-8B \citep{qwen3}, a hybrid language model that
supports both native reasoning (``think'') and direct (``no-think'') decoding
from the same underlying weights. We evaluate four policy
families:

\begin{enumerate}
\item \textbf{Base reasoning policy.} The instruction-tuned model decoded in
think mode.
\item \textbf{Decision-token classifier.} A LoRA \citep{hu2021} fine-tuned model
trained to predict the intervention decision directly and decoded in no-think
mode.
\item \textbf{Masked supervised fine-tuning.} A reasoning-preserving objective
that optimizes only the final intervention decision while masking reasoning
tokens from the training loss.
\item \textbf{Reinforcement-learned reasoning policies.} Group Relative Policy
Optimization (GRPO) \citep{shao2024} beginning from the base instruction-tuned model in think
mode. We evaluate both an accuracy-only objective and an objective that
additionally rewards behavioral dependence on the model's own reasoning.
\end{enumerate}

To evaluate generality beyond the primary model, we additionally report
experiments on Qwen3-30B, Nemotron-3-Nano, GPT-4o, GPT-5.1, and Llama-3.2-3B.
Complete implementation details and hyperparameters are reported in Appendix~A.

\subsection{Faithfulness Evaluation}
We evaluate faithfulness using three complementary analyses. \emph{Representational probes} measure when intervention decisions become decodable
from hidden activations during reasoning. \emph{Behavioral interventions} compare
complete reasoning, truncated reasoning, and neutral filler to distinguish
dependence on reasoning content from dependence on the reasoning format itself.
\emph{Dual-probe analyses} compare internally represented intervention types with
the intervention types stated in the generated reasoning. These analyses are used
exclusively for evaluation and are never included in the optimization objectives.
Detailed probe validation, behavioral controls, and stated-versus-internal reason
analyses are provided in Appendices~D--F.

\subsection{Reinforcement Learning Reward}
The reinforcement-learning experiments optimize the final intervention decision
while preserving free-form reasoning generation. The accuracy objective rewards
only correct intervention decisions. The faithfulness objective augments this
reward with a behavioral dependence bonus,
\[
r = \mathbb{1}[\text{decision correct}]\,(1 + \lambda\,\mathrm{dep}),
\]
where $\mathrm{dep}$ indicates whether completing the model's own reasoning
changes the final decision relative to truncating that reasoning, and
$\lambda = 0.3$. Activation probes are never used as optimization targets. Complete optimization details are reported in
Appendix~A.

\section{Results}

\subsection{Capability and Auditability}
The strongest intervention policy and the most auditable policy are different
models. Direct decision training produces the highest deployment performance but
eliminates inspectable reasoning, whereas native reasoning yields a more
auditable policy at a measurable cost in decision quality. Table~\ref{tab:main}
summarizes this tradeoff across all evaluated policies.

\begin{table}[t]
\centering
\small
\setlength{\tabcolsep}{3pt}
\begin{tabular}{@{}llcccc@{}}
\toprule
Policy & Decode & F1 & FIR & MIR & AUROC \\
\midrule
Concurrent Work SFT$^{\dagger}$ & token & 0.740 & 0.044 & 0.521 & --- \\
GPT-4o zero-shot & no-think & 0.462 & 0.065 & 1.000 & --- \\
GPT-4o zero-shot & think & 0.501 & 0.060 & 0.938 & --- \\
GPT-5.1 zero-shot & no-think & 0.436 & 0.473 & 0.312 & --- \\
GPT-5.1 zero-shot & think & 0.288 & 0.732 & 0.188 & --- \\
\midrule
Base Qwen3-8B & no-think & 0.096 & 0.989 & 0.026 & --- \\
Base Qwen3-8B & think & 0.536 & 0.232 & 0.579 & 0.637 \\
Token-SFT classifier & no-think & \textbf{0.620} & 0.276 & 0.079 & --- \\
Token-SFT classifier & think & 0.367 & 0.624 & 0.237 & --- \\
RL-accuracy & think & 0.520 & 0.329 & 0.395 & 0.640 \\
RL-faithfulness$^{\ddagger}$ & think & 0.549 & 0.254 & 0.474 & 0.622 \\
\midrule
\multicolumn{6}{@{}l}{\emph{Other base models}$^{\star}$} \\
Qwen3-30B & no-think & 0.104 & --- & --- & --- \\
Qwen3-30B & think & 0.494 & --- & --- & --- \\
Nemotron-3-Nano & no-think & 0.074 & --- & --- & --- \\
Nemotron-3-Nano & think & 0.420 & --- & --- & --- \\
Llama-3.2-3B$^{\S}$ & no-think & 0.000 & --- & --- & --- \\
Llama-3.2-3B$^{\S}$ & think & 0.084 & --- & --- & --- \\
\bottomrule
\end{tabular}
\caption{Decision quality across policies and models. Unmarked rows use the
matched evaluation ($n = 400$, seed 7). FIR: false-intervention rate; MIR:
missed-intervention rate. $^{\star}$Evaluated on a smaller 200-item slice drawn
from the same seed-7 shuffle, so absolute values are not directly comparable to
the rows above; where the two slices overlap they agree closely (Qwen3-8B
no-think 0.096 vs 0.097, think 0.536 in both). A Qwen3-30B token-SFT classifier
reaches 0.713 no-think, degrading to 0.562 in think mode.
$^{\dagger}$Llama-3.1-8B \citep{when2speak}. $^{\ddagger}$Reports the deployed
arm; the behavioral audits in Appendices~E--G use a separate checkpoint trained
under the same objective with a truncation-dependence counterfactual.
$^{\S}$Base, non-instruction-tuned, with no native reasoning toggle; it follows
the prompted-CoT format only 24\% of the time and is therefore used for probing
only.}
\label{tab:main}
\end{table}

\paragraph{Reasoning is essential for the base model.}
When decoded without reasoning, the base Qwen3-8B model predicts intervention for
nearly every example, producing a macro-F1 of 0.096 and a false-intervention rate
of 0.989. Decoding the same weights in think mode produces a policy that
discriminates rather than always intervening (macro-F1 0.536,
false-intervention rate 0.232). However,
missed interventions rise from 0.026 to 0.579. For the base model, intervention
timing is therefore expressed largely through its reasoning mode rather than
through direct prediction.

\paragraph{Decision-token fine-tuning produces the opposite behavior.}
Training only the final decision token yields the strongest direct classifier,
achieving a macro-F1 of 0.620 on the matched evaluation. However, decoding this
same checkpoint in think mode reduces performance to 0.367. The optimization that
improves direct classification also degrades the reasoning behavior
that makes the policy inspectable.

\paragraph{The tradeoff is not unique to a single model.}
Zero-shot frontier models exhibit similar limitations despite greater general capabilities. GPT-4o is conservative and misses most intervention
opportunities, whereas GPT-5.1 intervenes much more aggressively and incurs many
false interventions. Prompting either model to reason explicitly does not
reliably improve deployment performance, suggesting that intervention timing
under asymmetric costs is a learned policy rather than a capability that
consistently emerges through prompting.

The same qualitative behavior also replicates across model scale and family
(Table~\ref{tab:main}, lower block). Qwen3-30B, a four-times larger model in
the same family, and Nemotron-3-Nano, a different family with its own native
reasoning toggle, both perform poorly when decoded without reasoning but
improve in think mode. A decision-token classifier
trained on Qwen3-30B achieves strong no-think performance but again degrades when
evaluated in think mode. Although absolute performance varies across
architectures, the relationship between reasoning, capability, and auditability
remains consistent.

These results establish the central tradeoff investigated throughout the
remainder of the paper. If the objective is maximum deployment performance,
direct decision policies perform best but provide no reasoning to inspect. If the
objective is auditability, reasoning-mode policies expose the decision process but
incur a measurable performance cost.

\subsection{Supervised Fine-Tuning}
Straightforward supervised fine-tuning does not eliminate the
capability--auditability tradeoff. Decision-token supervision produces a stronger
direct classifier but suppresses reasoning, whereas a reasoning-preserving
objective maintains auditability without improving difficult intervention
decisions.

\paragraph{Decision-token fine-tuning suppresses reasoning.}
Fine-tuning the model directly on the intervention
decision improves no-think deployment performance
(Table~\ref{tab:main}), but it also degrades the reasoning behavior required for
auditability. Decoded in think mode, the same checkpoint reaches a macro-F1 of
only 0.367, below the untrained base policy's 0.536, and attempts to restore
reasoning through prompting or prefilling reduce deployment performance further
without recovering faithful reasoning. The optimization objective therefore
shifts the model away from the reasoning mode rather than improving it.

\paragraph{Masked supervised fine-tuning preserves reasoning but cannot improve
hard cases.}
We next consider a reasoning-preserving objective. During training, the model's
own reasoning trace is retained while the loss is applied only to the final
intervention decision. This masked objective preserves native reasoning
generation, with the model continuing to produce reasoning on approximately 99\%
of evaluation examples.

However, preserving reasoning alone does not improve deployment performance
beyond the base reasoning policy. The limitation follows from the construction of
the training data. To avoid answer-conditioned rationales, masked fine-tuning is
performed only on reasoning traces that the base model generated naturally and
that already produced the correct intervention decision. These examples represent
behaviors the model already performs successfully, providing little supervision
for confidently wrong decisions. Consequently, masked imitation reinforces
existing behavior without teaching the policy to solve difficult intervention
cases.

\begin{figure}[t]
\centering
\includegraphics[width=\columnwidth]{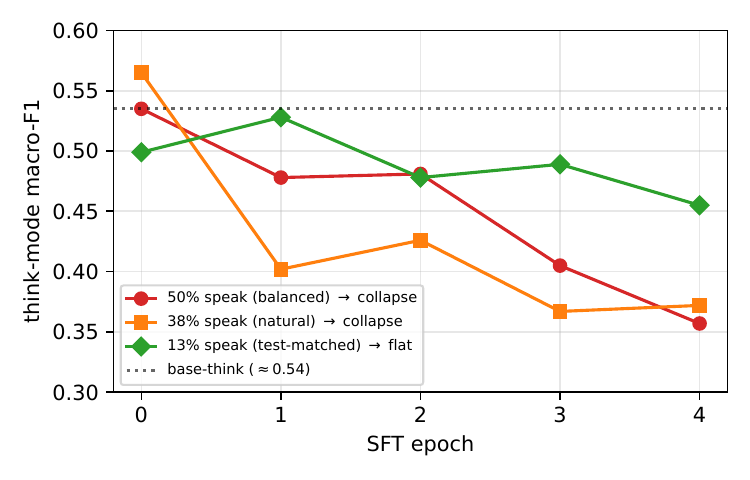}
\caption{The masked fine-tune degrades only when the intervention frequency in
the training data exceeds that of the deployment distribution. Think-mode
macro-F1 declines for intervention-enriched training mixtures but remains stable
when the training prior matches deployment. Reasoning generation and AUROC remain
essentially unchanged.}
\label{fig:dose}
\end{figure}

\paragraph{Apparent collapse is explained by prior mismatch.}
An apparent decline in masked-SFT performance occurs only when the intervention
frequency in the training data exceeds that of the deployment distribution. When
the training mixture matches the deployment prior, this degradation is reduced. Moreover, reasoning generation and threshold-free
discrimination remain essentially unchanged across training mixtures, indicating
that the observed decline reflects a shift in the operating point rather than a
loss of reasoning ability or class discrimination (Figure~\ref{fig:dose};
Appendix~B).

Together, these results show that supervised fine-tuning cannot recover both
capability and auditability. Optimizing only the decision token suppresses
reasoning, while preserving reasoning through masked supervision leaves the model
unable to improve.

\subsection{Reinforcement Learning}
We next ask whether reinforcement learning can improve the auditable reasoning
policy without relying on \emph{post hoc} rationales. Starting from the base
think policy, we train with GRPO under both accuracy-only and faithfulness-aware
rewards. Neither substantially improves deployed performance. The base policy
achieves a macro-F1 of 0.536, compared with 0.520 and 0.549 for the two RL
variants, while AUROC remains statistically similar across all policies. These
differences fall within the between-run variability we measure by repeating an
objective across independent random seeds (Appendix~G), so we treat them as
indistinguishable rather than as an ordering. Group-relative RL therefore
provides no meaningful improvement over the base reasoning policy.

One possible explanation is insufficient optimization. To test this possibility,
we increased rollout diversity, usable gradient, and training
duration. Despite an estimated 5--10$\times$ increase in effective gradient,
held-out performance plateaued at the same level (Appendix~C), suggesting that
the observed limitation is unlikely to result from undertraining.

\subsection{Decision Formation During Reasoning}
The reasoning policy forms its intervention decision during the chain of thought,
whereas the direct classifier commits before reasoning begins.

To examine when intervention decisions become represented, we train linear probes
on hidden activations before reasoning begins and after it ends. Under the
validation controls described in Appendix~D, including within-family evaluation,
text-only baselines, and nested layer selection, the base reasoning policy shows
little evidence that the intervention decision is represented before reasoning
beyond what is predictable from the input text. Pre-CoT activations achieve an
AUROC of $0.664 \pm 0.034$, compared with $0.631 \pm 0.053$ for a text-only
baseline. After reasoning, decision decodability rises sharply, reaching an AUROC
of $0.976 \pm 0.045$.

The direct classifier exhibits a fundamentally different representational
pattern. Probed at the same prompt position in no-think mode, its intervention
decision is already highly decodable before any reasoning is generated (AUROC
0.997), exceeding the corresponding text baseline (0.621). This is
the expected signature of a direct decision policy: the intervention decision is
formed immediately and then emitted.

We also probe intervention type rather than intervention timing, where a five-way
probe asks whether the activations encode which kind of epistemic gap the
conversation contains. Type information is most strongly represented before
reasoning (scored by balanced accuracy, so chance is 0.20) and gradually weakens
over the chain of thought, suggesting that the model first identifies the
underlying epistemic gap before arriving at the final binary intervention
decision.

\begin{figure*}[t]
\centering
\includegraphics[width=0.92\textwidth]{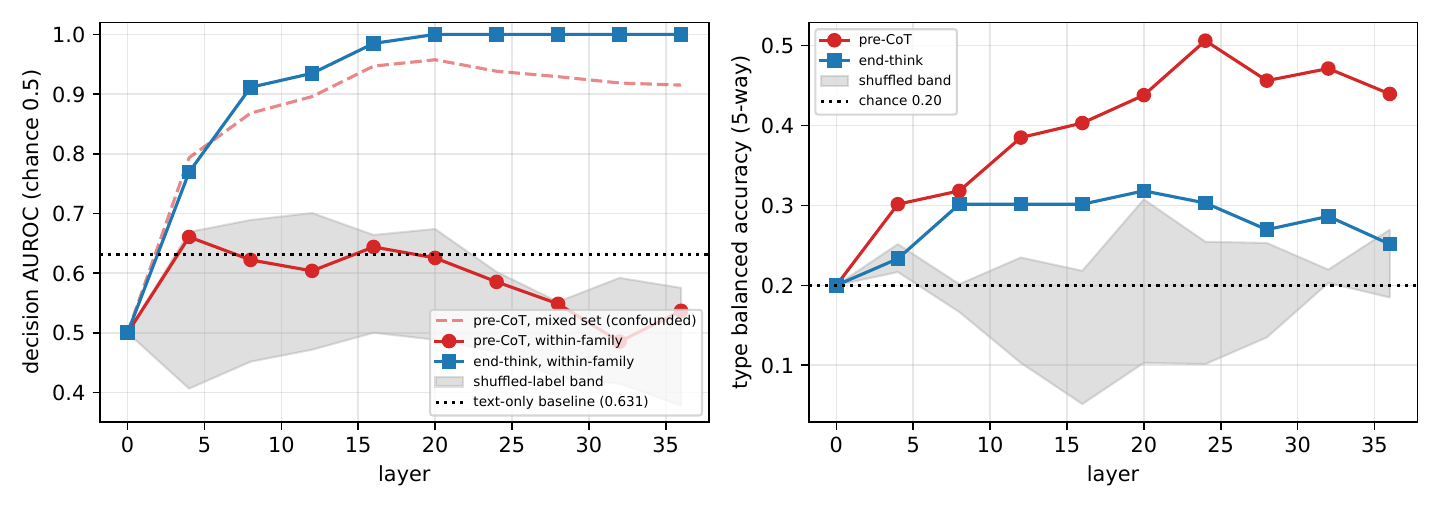}
\caption{Layer-resolved decision and intervention-type decodability for the base
Qwen3-8B reasoning policy. Each point is a linear probe trained on hidden
activations from one transformer layer (x-axis); the y-axis is held-out probe
performance (left: decision AUROC, chance 0.5; right: five-way type balanced
accuracy, chance 0.20). The gray band shows probes trained on shuffled labels.
Decision information remains close to the text baseline before reasoning and
becomes highly decodable only after reasoning, while intervention-type
representations are strongest before reasoning and weaken toward its conclusion.}
\label{fig:layers}
\end{figure*}

Figure~\ref{fig:layers} summarizes these dynamics across transformer depth.
Decision decodability remains close to the text baseline through most of the
reasoning process before increasing sharply at the end of the chain of thought,
whereas intervention-type information follows the opposite trajectory. These
findings suggest that the reasoning policy does not simply verbalize a
pre-existing intervention decision. Instead, the intervention decision appears to
develop during the reasoning process, whereas direct classifiers commit to the
action before any explanation is produced.

\subsection{Behavioral Dependence on Reasoning}
Behavioral interventions indicate that reasoning content has a measurable, but
modest, causal influence on intervention decisions once reasoning-format effects
are controlled.

Representational probes show when intervention decisions become decodable but
cannot establish whether those representations causally influence behavior. We
therefore complement the probe analyses with behavioral interventions that
compare complete reasoning, truncated reasoning, and neutral filler.

Removing reasoning entirely causes the base model to degenerate into an almost
always-intervene policy, demonstrating that reasoning mode is necessary for
usable intervention behavior. The more specific question is whether the
\emph{content} of the reasoning contributes to the final decision. A naive
chain-of-thought truncation analysis suggests dependence: truncating
reasoning changes many intervention decisions. However, replacing the removed
reasoning with length-matched neutral filler changes even more decisions. Thus, disrupting the reasoning format alone pushes
the model toward its no-think policy.

To isolate the contribution of reasoning content, we compare truncated reasoning
with the neutral-filler control. Retaining the first half of the original
reasoning consistently pulls predictions back toward the full-reasoning policy,
demonstrating that reasoning content contributes beyond the reasoning format
itself. For the base reasoning policy, this content-dependent effect is modest but
statistically reliable (Appendix~E). Across reinforcement-learning policies, the
same qualitative behavior persists, although the magnitude varies across independent training runs.

These results indicate that the generated reasoning is neither fully causal nor
simply decorative. Raw chain-of-thought ablations overestimate reasoning
dependence because they confound reasoning content with reasoning format, but
after controlling for this artifact, the reasoning itself retains a measurable
influence on the model's intervention decisions.

\subsection{Stated and Internal Reasons}
Generated explanations become more consistent with the gold intervention type
after reinforcement learning, but this does not imply that they faithfully
reflect the model's internal reasoning. The intervention type \emph{stated} in
each generated reasoning trace is extracted automatically by an LLM classifier
into the same five categories and human validation of this annotation is reported in
Appendix~F.

Reinforcement learning modestly increases agreement between the stated
intervention type and the gold intervention type, rising from 0.378 for the base
policy to 0.425 for the accuracy objective and 0.474 for the faithfulness
objective. Thus, reinforcement learning produces explanations that more
frequently describe the correct intervention category.

However, the probe analyses reveal an important qualification. At the end of the
chain of thought, internal representations align more closely with
the stated intervention type than with the gold intervention type. By this point,
however, the residual stream already contains the generated explanation, so the
probe may simply recover information from the text the model has just produced
rather than an internal representation that existed beforehand.

Before reasoning begins, this alignment largely disappears across all policies
(Appendix~F). This suggests that the stated explanation is constructed during the
reasoning process rather than read directly from a pre-existing internal
representation. More broadly, it highlights a limitation of dual-probe analyses:
probes applied after reasoning may recover the generated explanation itself,
whereas probes before reasoning may precede the emergence of an explicit reason
representation. Increasing agreement between generated explanations and gold
labels should therefore not be interpreted as evidence of increased reasoning
faithfulness.

\subsection{Behavioral Faithfulness Rewards Do Not Improve Faithfulness}
Adding a behavioral faithfulness reward does not measurably improve the reasoning
policy beyond the base model or an accuracy-only reward (Table~\ref{tab:main}). Across all held-out
audits, the faithfulness objective performs similarly to the accuracy objective
and the base reasoning policy. Deployment performance, decision-timing probes,
pre-CoT stated-reason alignment, and filler-controlled behavioral dependence
show little change. Importantly, the targeted behavioral
quantity itself does not improve after optimization.

One explanation follows from the behavioral analyses above. The reward is
computed from raw reasoning dependence, but raw chain-of-thought perturbations
are dominated by reasoning-format effects rather than reasoning content. As a
result, the policy can increase the reward by becoming dependent on the presence
of a reasoning span without becoming more dependent on the reasoning itself. A
behavioral reward defined without a content control therefore does not reliably
increase reasoning faithfulness.

Finally, reinforcement-learning objectives should be compared across independent
optimization runs rather than individual evaluation examples. Repeating the
accuracy objective across three random seeds revealed greater variation between
training runs than between reward objectives (Appendix~G). We therefore treat the
optimization run, rather than the evaluation item, as the appropriate unit of
inference when evaluating these objectives.

\section{Discussion}

Our results identify a practical boundary for explainable abstaining policies.
The strongest intervention policy is a direct classifier that commits to an
action before any explanation is produced, leaving nothing to audit. The most
auditable policy reasons explicitly before acting, but this transparency comes
with a measurable cost in deployment performance. Neither supervised fine-tuning
nor group-relative reinforcement learning closes this gap.

These findings suggest that requiring a model to explain an abstaining decision
changes both the optimization problem and the evaluation problem. Unlike question
answering, where explanations justify an already-required answer, abstaining
policies must explain why an action should occur at all. This distinction makes
calibration, class imbalance, and reasoning mode first-order components of
faithfulness rather than implementation details.

\subsection{Methodological Implications}
The experiments also identify several methodological considerations for
evaluating reasoning faithfulness in action policies.

\emph{First, probability-based contribution metrics fail on low-entropy actions.}
Stepwise metrics that measure how much each reasoning step raises the probability
of the final answer are informative for multi-token answers \citep{zhao2025tts},
but here the decision token's probability is near one whether reasoning is
present, absent, or corrupted. The metric saturates at zero contribution and
cannot separate genuine reasoning from known post-hoc controls.

\emph{Second, calibration should be evaluated separately from discrimination.}
Both class-balanced supervised learning and reinforcement learning altered the
intervention threshold without substantially changing threshold-free
discrimination. Reporting only deployment metrics can therefore confuse
operating-point shifts with genuine improvements or degradations in
representational capability.

\emph{Third, probe methodology requires stronger controls than are typically used
for question-answering tasks.} Our probe analyses show that raw probe accuracy
becomes misleading under severe class imbalance, mixed evaluation sets introduce
construction confounds, and probe parameters do not necessarily transfer across
evaluation registers. These observations reinforce previous cautions that probe
performance alone is not sufficient evidence of internal representations or
causal computation \citep{hewitt2019,burns2022,hiddenerror}.

\emph{Fourth, probe position fundamentally changes interpretation.}
End-of-reasoning probes may recover information directly from the generated
explanation rather than from the internal computation that produced it, whereas
probes before reasoning may precede the emergence of an explicit reason
representation.

\emph{Fifth, behavioral perturbations require content controls.} Raw
chain-of-thought truncation overestimates reasoning dependence
because disrupting the reasoning format alone pushes the model toward its
degenerate no-think policy. Comparing truncation with length-matched neutral
filler isolates the contribution of reasoning content.

\emph{Sixth, reinforcement-learning objectives should be compared across
independent optimization runs rather than individual evaluation examples.} Our
repeated training runs showed that between-run variability exceeded the observed
differences between reward objectives. This recommendation extends beyond GRPO
and applies equally to stochastic RL and
preference-optimization pipelines.

\emph{Finally, behavioral faithfulness rewards require careful construction.} Our
reward was intentionally restricted to behavioral signals rather than activation
probes to avoid directly optimizing the audit itself. However, the results
suggest that behavioral rewards derived from uncontrolled perturbations may
primarily capture dependence on reasoning format rather than dependence on
reasoning content. This illustrates a broader form of Goodhart's law: optimizing
a behavioral proxy for faithfulness does not necessarily improve the underlying
reasoning process \citep{skalse2022}.

\subsection{Limitations and Future Work}
Several limitations remain. The experiments focus on a single intervention-timing
benchmark composed of synthetic conversations. Although the
capability--auditability tradeoff replicates across model scale and family,
complete probe and reinforcement-learning analyses are limited to Qwen3-8B.

Our representational analyses also remain correlational. Although the behavioral
interventions provide complementary causal evidence, stronger causal techniques,
including activation steering, causal mediation analysis, and circuit-level
interventions, could more directly test whether the identified representations
are necessary.

Several algorithmic directions also emerge naturally from these results. The
masked supervised objective motivates multi-round self-distillation under
deployment-matched class priors, while the reinforcement-learning experiments
suggest exploring objectives that provide learning signals even when sampled
rollouts unanimously select the wrong action, such as approaches based on learned
value functions or global baselines. Future work should also investigate rewards
derived from content-controlled perturbations and explore optimization objectives
that improve independent behavioral and representational audits simultaneously.

\section{Conclusion}
We studied faithful reasoning in an abstaining action policy using intervention timing as a controlled testbed. We find a consistent capability–auditability tradeoff: direct classification yields stronger intervention decisions, while explicit reasoning provides greater auditability at the cost of performance. Neither supervised fine-tuning nor group-relative reinforcement learning closes this gap. At the same time, controlled probes and behavioral interventions provide evidence that reasoning contributes to decision formation rather than merely rationalizing a predetermined action. Together, these findings show that faithfulness methods developed for question answering do not transfer directly to action policies and provide a foundation for studying faithful reasoning in agentic systems where deciding whether to act is as important as deciding what to say.

\section{Reproducibility Statement}
All supervised fine-tuning uses LoRA adapters (rank 16, $\alpha = 32$, zero dropout, all linear layers), and reinforcement learning uses GRPO. Activation probes are L2-regularized logistic regressions trained on cached hidden states. Experiments run on a single 24 GB GPU. We report three random seeds per reinforcement-learning objective, use exact McNemar tests for paired behavioral interventions and bootstrap confidence intervals where applicable, and evaluate models at their natural operating points without post hoc threshold tuning. Detailed methods, hyperparameters, controls, and reproducibility procedures are provided in the Appendix. Upon publication, we will open-source the code, fine-tuned adapters, prompts, intervention mappings, evaluation scripts, and the methodological controls required to reproduce all reported results.

\bibliography{refs}

\newpage
\section{Appendix}
\noindent
This supplement provides the experimental details, controls, and complete
analyses supporting the main paper. Table, figure, and section numbers prefixed
with a letter refer to this appendix; unprefixed numbers refer to the main
paper.

\section{Appendix A. Experimental Details}

\subsection{A.1 Dataset}
We evaluate all methods on the When2Speak intervention-timing benchmark
\citep{when2speak}. The task is binary classification: given a prefix of a
multi-party conversation, the model predicts whether an assistant should
intervene now or remain silent. Intervention opportunities comprise
approximately 13\% of decision points, resulting in a severely class-imbalanced
evaluation setting. Table~\ref{tab:a-data} summarizes the dataset statistics
used throughout the paper.

\begin{table}[h]
\centering
\small
\begin{tabular}{@{}lr@{}}
\toprule
Statistic & Value \\
\midrule
Source scenarios & 11,498 \\
Synthetic conversations & $\approx$16,000 \\
Token-level decision points & 173,325 \\
Intervention prevalence & $\approx$13\% \\
Intervention categories & 5 \\
\bottomrule
\end{tabular}
\caption{Dataset statistics.}
\label{tab:a-data}
\end{table}

\subsection{A.2 Prompt Templates}
All models receive identical task instructions except for the requested decoding
mode.

\begin{quote}\small\ttfamily
\noindent You are an epistemic facilitator in a multi-agent discussion. Decide
whether to intervene now. Output exactly one token: $<$ to speak, or $>$ to stay
silent.
\end{quote}

\noindent
No-think and think conditions use this same instruction and differ only in the
chat template's native reasoning toggle (\texttt{enable\_thinking} false or
true). For all think-mode evaluations, only the final decision token, read after
the reasoning span closes, is scored.

\subsection{A.3 Intervention-Type Reconstruction}
The released When2Speak token-level dataset does not include intervention-type
labels for individual decision points. We reconstruct these labels by mapping
each generated conversation back to its originating source scenario using the
inherited source identifiers. To validate the reconstruction, we compare
content-word overlap between linked conversation--source pairs and randomly
shuffled pairs. Correct mappings exhibit substantially greater overlap (0.694
versus 0.100), supporting the recovered labels. Recovered intervention types are
used only for probe analyses and stated-versus-internal reason comparisons. They
are never used to supervise intervention prediction.

\subsection{A.4 Model Configurations}
All primary experiments use Qwen3-8B \citep{qwen3} because the same checkpoint
supports both native reasoning (``think'') and direct (``no-think'') decoding.
Table~\ref{tab:a-modelcfg} summarizes every model evaluated in the paper.

\begin{table}[h]
\centering
\small
\begin{tabular}{@{}llp{2.4cm}@{}}
\toprule
Model & Parameters & Role \\
\midrule
Qwen3-8B & 8B & Primary experiments \\
Qwen3-30B & 30B & Same-family scaling \\
Nemotron-3-Nano-30B-A3B & 30B (MoE) & Cross-family replication \\
Llama-3.2-3B & 3B & Probe analyses only \\
GPT-4o & Proprietary & Frontier baseline \\
GPT-5.1 & Proprietary & Frontier baseline \\
\bottomrule
\end{tabular}
\caption{Model configurations.}
\label{tab:a-modelcfg}
\end{table}

\subsection{A.5 Supervised Fine-Tuning}
Decision-token supervised fine-tuning optimizes only the final intervention
decision using LoRA adapters \citep{hu2021}. Masked supervised fine-tuning
preserves the model's generated reasoning while masking reasoning tokens from the
training loss; optimization is again applied only to the final intervention
decision. To avoid training on post-hoc explanations, masked SFT uses only
reasoning traces that (i) were naturally generated by the base model before
supervision, and (ii) produced the correct intervention decision. Incorrect
reasoning traces are excluded because conditioning explanations on the known
correct answer would produce answer-conditioned rationales that are post hoc by
construction.

\begin{table}[h]
\centering
\small
\begin{tabular}{@{}lcc@{}}
\toprule
Hyperparameter & Token-SFT & Masked SFT \\
\midrule
LoRA rank & 16 & 16 \\
Learning rate & 1e-4 & 5e-5 \\
Batch size & 32 & 16 \\
Epochs & 2 & 4 \\
Optimizer & Adam & Adam \\
Max sequence length & 4096 & 2048 \\
\bottomrule
\end{tabular}
\caption{Supervised fine-tuning hyperparameters.}
\label{tab:a-sft}
\end{table}

\subsection{A.6 Reinforcement Learning}
All reinforcement-learning experiments begin from the instruction-tuned Qwen3-8B
checkpoint and operate exclusively in think mode. We use Group Relative Policy
Optimization \citep{shao2024}. Two reward functions are evaluated:
\[
r_{\text{acc}} = \mathbb{1}[\text{decision correct}],
\]
\[
r_{\text{faith}} = \mathbb{1}[\text{decision correct}]\cdot(1 + \lambda \cdot \text{dep}),
\]
where $\text{dep}$ is 1 when completing the model's own reasoning changes the
decision relative to truncating it partway, and $\lambda = 0.3$. Behavioral
probes, activation probes, and probe outputs are never used as optimization
targets.

\begin{table}[h]
\centering
\small
\begin{tabular}{@{}lc@{}}
\toprule
Hyperparameter & Value \\
\midrule
Algorithm & GRPO \\
Learning rate & 2e-5 \\
Rollout group size & 8 \\
Sampling temperature & 0.8 \\
$\lambda$ & 0.3 \\
LoRA rank & 16 \\
Prompts per iteration & 16 \\
Max rollout tokens & 1024 \\
Training iterations & 150 \\
\bottomrule
\end{tabular}
\caption{Reinforcement-learning hyperparameters. Hyperparameters for the
increased-gradient experiment are given separately in Appendix~C.}
\label{tab:a-rl}
\end{table}

\subsection{A.7 Probe Training}
Linear probes are trained independently for intervention decision, intervention
type, and stated intervention type. Each probe is an L2-regularized logistic
regression trained on hidden activations extracted from a single transformer
layer and token position. To reduce common sources of probe overestimation,
every reported probe incorporates the following controls: within-family
evaluation, TF-IDF text-only baselines, shuffled-label controls, and nested
cross-validation for layer selection. All probes use $C = 0.1$ with the lbfgs
solver ($C = 1.0$ for the TF-IDF text baselines), trained on 70/30 stratified
splits with every second layer probed. Shuffled-label controls are averaged over
five random seeds, and layer selection is nested inside the training split so
test data never informs it. Complete validation experiments are reported in
Appendix~D.

\subsection{A.8 Evaluation Metrics}
The primary deployment metric is macro-F1, which weights intervention and
non-intervention performance equally despite severe class imbalance. We
additionally report the false-intervention rate and the missed-intervention rate,
and use AUROC solely as a threshold-free diagnostic of class separability.

\subsection{A.9 Statistical Analysis}
Confidence intervals are estimated using bootstrap resampling (1,000 resamples)
of evaluation examples. Behavioral perturbation experiments use exact McNemar
tests because paired interventions are evaluated on identical examples. Probe
agreement comparisons in Appendix~F are tested against permutation nulls
constructed by shuffling labels (2,000 draws). For reinforcement-learning
experiments, independent training runs, not individual evaluation examples, are
treated as the unit of inference when comparing reward objectives. This avoids
overstating objective-level conclusions from stochastic optimization.

\subsection{A.10 Compute}
Training and evaluation were performed on NVIDIA GPUs.

\begin{table}[h]
\centering
\small
\begin{tabular}{@{}lp{4.2cm}@{}}
\toprule
Resource & Value \\
\midrule
GPU & NVIDIA A10G \\
GPU memory & 24 GB \\
Total training time & $\approx$2--3 h per RL run;
                      $\approx$1--2 h per SFT run (hosted platform) \\
Probe training time & $\approx$1--2 h per policy \\
Software & Python 3.12, PyTorch 2.13.0, scikit-learn 1.9.0, CUDA 13.0 \\
\bottomrule
\end{tabular}
\caption{Computational resources.}
\label{tab:a-compute}
\end{table}

\subsection{A.11 Experiment Summary}
Table~\ref{tab:a-summary} summarizes every experiment reported in the paper.

\begin{table*}

\centering
\small
\setlength{\tabcolsep}{3pt}
\begin{tabular}{@{}llll@{}}
\toprule
Experiment & Training & Evaluation & Reported in \\
\midrule
Capability comparison & None & Macro-F1 & Table 1 \\
Cross-model replication & None & Macro-F1 & Table 1 \\
Token SFT & LoRA & Macro-F1 & Supervised FT \\
Masked SFT & LoRA & Macro-F1 & Figure 1 \\
Prior-shift study & Masked SFT & Macro-F1, AUROC & Figure 1 \\
RL (accuracy) & GRPO & Macro-F1 & Reinforcement L. \\
RL (faithfulness) & GRPO & Macro-F1 & Reinforcement L. \\
RL optimization study & GRPO & Macro-F1 & Appendix C \\
Decision probes & Linear probes & AUROC & Figure 2 \\
Behavioral dependence & Ablations & Flip analysis & Behav.\ dependence \\
Stated vs.\ internal & Dual probes & Agreement & Stated/internal \\
Reward ablation & RL & Behavioral audits & Reward ablation \\
\bottomrule
\end{tabular}
\caption{Summary of experimental conditions. All experiments use Qwen3-8B except
the cross-model replication, which additionally evaluates Qwen3-30B,
Nemotron-3-Nano, and Llama-3.2-3B.}
\label{tab:a-summary}
\end{table*}

\section{Appendix B. Prior Shift Explains the Apparent Collapse of Masked
Supervised Fine-Tuning}

The main paper shows that masked supervised fine-tuning preserves reasoning
while appearing to reduce deployment performance when trained on
intervention-enriched datasets (Figure 1). This appendix provides additional
evidence that the observed degradation is explained by a mismatch between the
training and deployment class distributions rather than by a loss of reasoning
ability or class discrimination.

\subsection{B.1 Experimental Design}
The deployment distribution used throughout the paper contains intervention
opportunities in approximately 13\% of evaluation examples. To isolate the effect
of the training class prior, we construct three masked-SFT training mixtures that
vary only in intervention prevalence while leaving the evaluation distribution
unchanged.

\begin{table}[h]
\centering
\small
\begin{tabular}{@{}lcp{3.1cm}@{}}
\toprule
Training mixture & Speak prev. & Purpose \\
\midrule
Matched & 13\% & Matches deployment distribution \\
Moderately enriched & 38\% & Intermediate prior shift \\
Strongly enriched & 50\% & Balanced training mixture \\
\bottomrule
\end{tabular}
\caption{Training mixtures used in the prior-shift experiment.}
\label{tab:b-mixtures}
\end{table}

All models are initialized from the same instruction-tuned checkpoint and trained
using identical optimization settings. The only experimental variable is the
proportion of intervention examples presented during training.

\subsection{B.2 Reasoning Behavior Remains Preserved}
A possible explanation for the performance decline is that masked fine-tuning
gradually suppresses reasoning. We evaluate this directly by measuring the
fraction of evaluation examples for which the model generates a complete
reasoning trace. Reasoning remains essentially unchanged across all training
mixtures throughout optimization.

\begin{table}[h]
\centering
\small
\begin{tabular}{@{}lc@{}}
\toprule
Training mixture & Think rate \\
\midrule
Base model & $\approx$1.00 \\
13\% speak & $\approx$0.99 \\
38\% speak & $\approx$0.99 \\
50\% speak & $\approx$0.99 \\
\bottomrule
\end{tabular}
\caption{Think-mode generation rate during masked supervised fine-tuning.}
\label{tab:b-think}
\end{table}

Across all conditions, approximately 99\% of evaluation examples continue to
produce native reasoning, indicating that the masked objective preserves
reasoning behavior irrespective of class balance.

\subsection{B.3 Threshold-Free Discrimination Remains Stable}
If masked supervised fine-tuning were damaging the model's underlying ability to
distinguish intervention opportunities, threshold-free discrimination would also
decline. Instead, AUROC remains statistically unchanged at the two measured
endpoints.

\begin{table}[h]
\centering
\small
\begin{tabular}{@{}lcc@{}}
\toprule
Training mixture & Macro-F1 & AUROC \\
\midrule
Base model & 0.536 & 0.637 \\
13\% speak & 0.455 & --- \\
38\% speak & 0.372 & --- \\
50\% speak & 0.357 & 0.66 [0.58, 0.74] \\
\bottomrule
\end{tabular}
\caption{Threshold-free discrimination after masked supervised fine-tuning.
AUROC was measured for the base model and the most-shifted (50\%) mixture; the
stability claim rests on that endpoint comparison.}
\label{tab:b-auroc}
\end{table}

Although macro-F1 decreases substantially as intervention prevalence increases
during training, AUROC remains within overlapping confidence intervals. This
indicates that the model continues to rank intervention opportunities similarly,
while shifting the decision threshold used to convert scores into binary actions.

\subsection{B.4 Cross-Family Replication}
To determine whether the observed behavior depends on Qwen3-8B, we repeat the
experiment using Nemotron-3-Nano. The same qualitative behavior emerges. Masked
supervised fine-tuning preserves reasoning generation throughout training while
intervention-enriched mixtures again reduce deployment macro-F1 despite stable
threshold-free discrimination.

\begin{table}[h]
\centering
\small
\begin{tabular}{@{}lccc@{}}
\toprule
Model & Mixture & Think rate & Macro-F1 \\
\midrule
Base & --- & $\approx$1.00 & 0.420 \\
Masked SFT & 52\% speak & $\approx$0.99 & 0.26$^{\dagger}$ \\
\bottomrule
\end{tabular}
\caption{Cross-family replication on Nemotron-3-Nano. $^{\dagger}$From 0.44 at
epoch 0.}
\label{tab:b-cross}
\end{table}

Although absolute performance differs from Qwen3-8B, the qualitative pattern is
unchanged. Increasing intervention prevalence during training primarily shifts
the deployed operating point rather than reducing the model's ability to
distinguish intervention opportunities.

\subsection{B.5 Interpretation}
Taken together, these experiments support the interpretation presented in the
main paper. Three independent observations remain consistent across training
mixtures: reasoning generation remains essentially unchanged; AUROC remains
stable; and deployment macro-F1 decreases only when the training intervention
prior exceeds the deployment prior. These observations are difficult to reconcile
with an explanation based on degraded reasoning or reduced representational
capacity. Instead, they indicate that masked supervised fine-tuning primarily
changes the model's calibration. Training on intervention-enriched data shifts
the learned decision threshold toward more frequent intervention, increasing
false-positive interventions when evaluated under the naturally imbalanced
deployment distribution. Consequently, the apparent collapse shown in Figure 1 is
best understood as an operating-point shift induced by prior mismatch rather than
a loss of reasoning ability or class discrimination.

\section{Appendix C. Increasing RL Signal Does Not Eliminate the Performance
Plateau}

\subsection{C.1 Motivation}
A natural alternative to supervised fine-tuning is reinforcement learning, which
allows the model to generate its own reasoning and receive reward based on the
correctness of its final intervention decision. Because RL does not require
answer-conditioned rationales, it avoids training on explanations that are post
hoc by construction. Our primary experiments found no meaningful improvement in
deployed performance over the base think policy. One possible explanation is that
the RL runs were simply undertrained. We therefore conducted an additional
experiment to determine whether substantially increasing the amount of usable
optimization signal improves performance.

\subsection{C.2 Group-Relative RL Can Produce Little Learning Signal}
Group-relative policy optimization computes rewards relative to other sampled
rollouts from the same prompt. When all sampled rollouts receive the same reward,
every rollout has zero centered advantage and no policy update is produced. This
situation commonly occurs when the model is confidently correct or confidently
incorrect. In particular, prompts for which every rollout predicts the same
incorrect intervention decision contribute little or no gradient, even though
they represent the examples where additional learning would be most valuable.

\subsection{C.3 Increasing the Amount of Usable Gradient}
To determine whether this optimization behavior explained the observed plateau,
we substantially increased rollout diversity. Compared with the original RL
configuration, the pushed run increased sampling temperature from 0.8 to 1.0,
doubled the rollout group size from 8 to 16, and increased training from 150 to
250 iterations. These changes reduced the proportion of zero-variance rollout
groups from approximately 15/16 groups to approximately 3/16 groups,
corresponding to roughly a five- to ten-fold increase in usable policy gradient.

\subsection{C.4 Results}
\begin{figure}[h]
\centering
\includegraphics[width=\columnwidth]{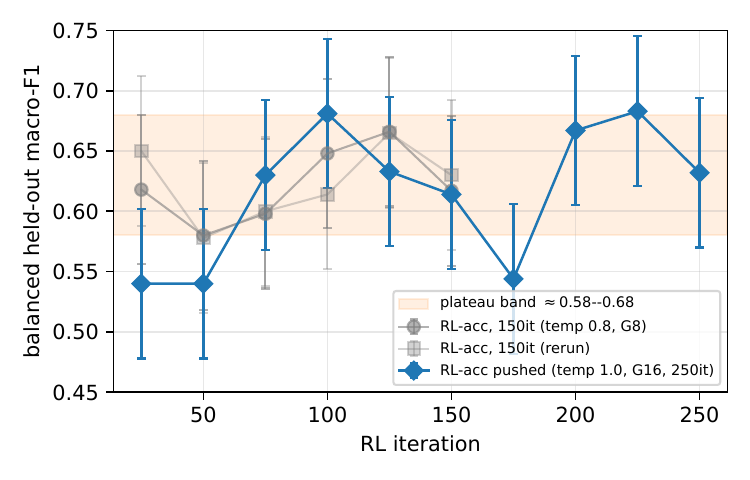}
\caption{Held-out macro-F1 throughout reinforcement learning. Increasing rollout
diversity (temperature 1.0, groups of 16, 250 iterations) reduced zero-variance
rollout groups by approximately five- to ten-fold but did not produce a higher
performance plateau than the original configuration. Error bars show $\pm$1 SE
across the 60-item held-out evaluation set.}
\label{fig:c-rl}
\end{figure}

Despite the substantially larger optimization signal, held-out performance
remained within essentially the same range throughout training
(Figure~\ref{fig:c-rl}). The pushed configuration briefly reached similar peak
macro-F1 values but did not establish a higher performance plateau than the
original runs. These results indicate that the lack of improvement is unlikely to
be explained by insufficient optimization. Instead, increasing both rollout
diversity and effective gradient leaves the achievable performance essentially
unchanged.

\subsection{C.5 Interpretation}
This behavior mirrors the limitation observed under supervised fine-tuning.
Imitation learning only receives supervision on behaviors represented in the
training targets, while group-relative RL only receives a learning signal when
sampled rollouts disagree. Confident errors therefore remain difficult for both
approaches to correct: supervised learning cannot teach behaviors absent from the
demonstrations, and GRPO receives no update when every sampled rollout makes the
same mistake. Alternative RL algorithms that employ learned value functions or
global baselines could, in principle, provide non-zero learning signals on
uniformly incorrect groups. Whether such methods improve intervention timing
while preserving reasoning faithfulness remains an open question.

\subsection{C.6 Reaching the Deployed Operating Point}
When2Speak's own reinforcement-learning stage uses asymmetric reward shaping to
cut the missed-intervention rate to approximately 0.19 to 0.22, on a policy with
no reasoning channel \citep{when2speak}. Because operating-point shaping is
orthogonal to auditability, we test whether the auditable policy can be moved the
same way: an RL arm trained from the base model in think mode with an asymmetric
reward (correct $+1$, false intervention $0$, missed intervention $-0.5$). In a
single run at one penalty ratio, scoped as a within-run existence claim
consistent with the seed-variance caution of Appendix~G, the missed-intervention
rate fell to 0.158, with speak recall 0.84 and a think rate of 1.00. The
auditable channel therefore survives operating-point shaping intact. At this
untuned setting the cost was severe over-intervention (false-intervention rate
0.790, macro-F1 0.261); reaching a balanced configuration is a tuning exercise
rather than an auditability barrier. This extends the conclusion of this
appendix: reward shaping moves the policy along its error-tradeoff frontier,
while the ceiling documented above concerns the height of that frontier, which
shaping does not raise.

\section{Appendix D. Probe Validation}

The main paper uses linear probes to study when intervention decisions become
represented during reasoning. Because probe analyses are sensitive to evaluation
methodology, this appendix summarizes the controls used to distinguish genuine
decision representations from artifacts of dataset construction, class imbalance,
and probe selection.

\subsection{D.1 Probe Methodology}
All probes are L2-regularized logistic regressions trained independently for a
single transformer layer and token position. Decision probes predict the final
intervention decision, while type probes predict either the gold intervention
type or the intervention type stated in the model's generated reasoning. Probe
performance is evaluated on held-out examples using balanced accuracy and AUROC.
Because intervention opportunities comprise only approximately 13\% of the
evaluation set, AUROC is treated as the primary measure of decodability
throughout the paper.

\paragraph{Intervention-type probes.}
The type probes use the same activations and the same probe recipe as the
decision probes, with a different label. Each typed evaluation item carries a
gold intervention category recovered from its source scenario (approximately 40
items per category), and the category is a property of the input, known from the
dataset rather than from anything the model generates. A multinomial logistic
regression receives the hidden activation at a single layer and position and
predicts one of the five categories, trained on 70\% of items and scored on the
held-out remainder. Because the categories are not perfectly balanced, the probe
is scored with balanced accuracy (average per-class accuracy) so chance is 0.20
regardless of class frequencies. Probes trained on shuffled category labels reach
approximately 0.29, setting the effective floor.

Before reasoning begins, the type probe reaches a balanced accuracy of 0.506 from
the pre-CoT activation alone: the category is identified correctly about half the
time, five-way. By the end of reasoning this falls to 0.335. As the chain of
thought unfolds, decision-directed computation accumulates in the residual stream
and the type-specific input features partially wash out. Together with the
decision probes, this yields a consistent temporal picture: the model represents
what kind of situation it faces immediately upon reading the context, and spends
the reasoning converting that assessment into what to do. The type signal is
strongest early and fades, while the decision signal is absent early and appears
late. Decodability does not establish use; the behavioral analyses in Appendix~E
carry the causal evidence. And because the category is an input property, strong
pre-CoT type decodability primarily reflects input comprehension, which is also
why the presence of rich, type-correlated input features motivated the
within-family and text-baseline controls applied to the decision probes.

\subsection{D.2 Probe Accuracy Under Class Imbalance}
A naive analysis based on probe accuracy can underestimate representational
information under severe class imbalance. In the intervention-timing task,
predicting the majority class already produces high accuracy, making raw accuracy
an unreliable indicator of decodable information. Conversely, AUROC measures
whether probe outputs correctly rank intervention and non-intervention examples
independently of the decision threshold. For this reason, the main paper reports
AUROC as the primary probe metric and uses balanced accuracy only as a secondary
reference.

\subsection{D.3 Controlling for Dataset Construction}
Mixed evaluations containing both intervention examples and ordinary
conversational turns introduce a potential construction confound. Intervention
examples originate from reconstructed source scenarios, whereas many silent
examples originate from ordinary background dialogue. A probe trained across the
entire dataset may therefore distinguish conversation families rather than decode
an emerging intervention decision. To control for this possibility, we repeat the
analyses within individual intervention families, where lexical content and task
structure are substantially more similar. Under this within-family evaluation,
pre-reasoning decision decodability falls close to the text-only baseline, while
post-reasoning decodability remains near ceiling. The qualitative conclusion
reported in the main paper therefore does not depend on the mixed-set
construction.

\subsection{D.4 Text-Only Baselines}
To determine how much information is already present in the prompt itself, we
compare activation probes with a TF-IDF logistic-regression classifier trained
directly on the input text. Across pre-CoT analyses, activation probes exceed the
text baseline only modestly. After reasoning, however, probe performance
substantially exceeds the text baseline, indicating that additional decision
information becomes represented during reasoning rather than being recoverable
from the prompt alone. This comparison prevents interpreting predictable input
features as evidence that the intervention decision has already been internally
committed.

\subsection{D.5 Layer Selection}
Probe performance varies substantially across transformer layers. Selecting the
highest-scoring layer on the evaluation set would therefore produce optimistic
estimates of decodability. To avoid this bias, all reported layers are selected
using nested cross-validation. Layer selection is performed entirely within the
training folds, while the held-out evaluation split remains unseen until final
evaluation. No reported probe result uses the evaluation set for model selection.

\subsection{D.6 Register Transfer}
Most probe analyses use reconstructed three-line contexts derived from the
original source scenarios in order to isolate the intervention decision while
minimizing unrelated conversational variation. To test whether these findings
generalize to deployment conditions, we repeat the analyses on complete
multi-party conversations sampled from the evaluation set. The overall
representational pattern replicates: intervention type is represented before
reasoning and weakens toward the end of the chain of thought, while intervention
decisions become substantially more decodable after reasoning.

However, probe parameters do not transfer directly across registers. A probe
trained on reconstructed contexts performs near chance when evaluated on complete
conversations. This indicates that activation probes should be trained and
evaluated within the same representational register. The pre-CoT decision signal
is also register-dependent. On reconstructed contexts, pre-CoT decodability
remains close to the text baseline. On full conversations, a modest pre-CoT
signal beyond the text baseline is present, although the decision remains
substantially less decodable than after reasoning. We therefore interpret the
pre-CoT result as register-specific rather than universal.

\subsection{D.7 Probe Position Matters}
The interpretation of a probe depends critically on where activations are
extracted. Pre-CoT probes measure information available before reasoning begins.
Intermediate probes measure representations while reasoning unfolds. End-of-CoT
probes measure representations after the model has already generated its
explanation. This distinction is particularly important for the
stated-versus-internal reason analysis. Representations extracted after reasoning
may contain information copied directly from the generated explanation itself.
Consequently, agreement between end-of-CoT probes and the stated explanation is
insufficient evidence that the stated reason existed internally before reasoning
began.

\subsection{D.8 Decision Formation Along the Chain of Thought}
\begin{figure}[h]
\centering
\includegraphics[width=\columnwidth]{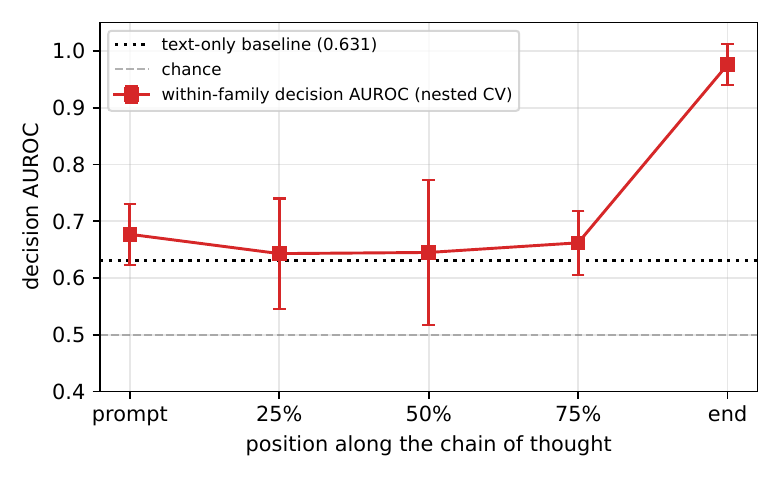}
\caption{Decision decodability along the chain of thought (base policy,
within-family decision AUROC, nested cross-validation, mean $\pm$ std over ten
splits). Decodability is flat at the text-baseline level through 75 percent of the
reasoning and rises to near-ceiling only at its close.}
\label{fig:d-traj}
\end{figure}

To locate when during reasoning the decision forms, we extract activations at
four points within the base policy's chain of thought (25, 50, and 75 percent of
the reasoning span, and its end) and probe the within-family decision at each
position using nested layer selection over ten splits.
Figure~\ref{fig:d-traj} shows the resulting trajectory. Decodability remains flat
at the text-baseline level through three-quarters of the reasoning span (AUROC
0.64 to 0.68, statistically indistinguishable from the 0.631 text-only baseline)
and rises to 0.976 only at the close of the span. The decision therefore
crystallizes late rather than emerging gradually. This is consistent with the
behavioral analyses of Appendix~E, since truncating at 50 percent cuts before the
typical commitment point, making re-decision from half a reasoning trace
genuinely counterfactual. Caveats: four sampled positions, $n = 196$, and
fractional positions correspond to different absolute depths across items.

\subsection{D.9 Layer-Resolved Probes per Policy}
\begin{figure*}[t]
\centering
\includegraphics[width=0.92\textwidth]{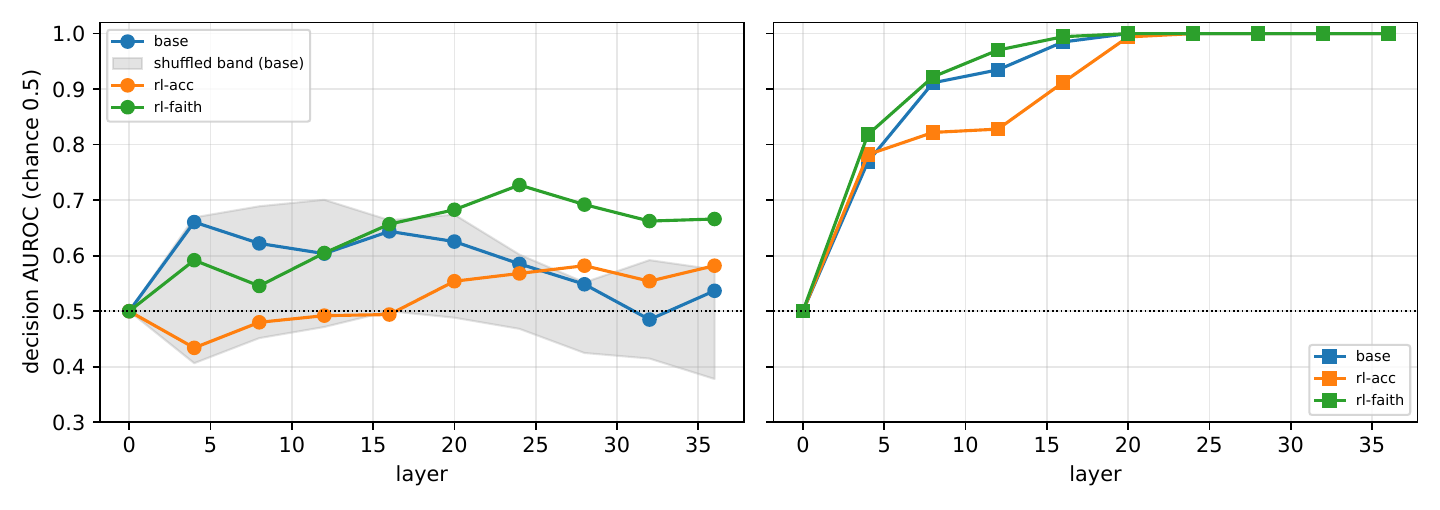}
\caption{Within-family decision decodability by layer for the base,
RL-accuracy, and RL-faithfulness policies (AUROC, chance 0.5; shaded region shows
the shuffled-label band computed on the base policy). Left panel: pre-CoT, all
policies within the noise band at all depths. Right panel: end-of-think, all
policies saturate together by roughly layer 16 to 20.}
\label{fig:d-arms}
\end{figure*}

Repeating the within-family timing analysis separately for the base policy and
both reinforcement-learning arms asks whether reinforcement learning moves where
in the network, or when relative to the chain of thought, the decision becomes
readable. It does not appear to (Figure~\ref{fig:d-arms}): before the chain of
thought, all three policies remain within or near the shuffled-label band at
every depth, and after the chain of thought all three rise together and saturate
by roughly layer 16 to 20. The reward objectives therefore differ neither
behaviorally (Appendix~G) nor in their decision-formation profile.

\subsection{D.10 Summary}
Taken together, these controls support the interpretation of the probe analyses
presented in the main paper. Specifically, AUROC provides a more reliable measure
of probe performance than raw accuracy under severe class imbalance; mixed-set
evaluations overestimate pre-reasoning decision information because they
introduce dataset construction confounds; text-only baselines explain much of the
apparent pre-CoT signal; nested cross-validation prevents optimistic layer
selection; probe parameters do not transfer reliably across evaluation registers;
and probe position fundamentally changes what internal representations can be
interpreted. These controls support the conclusion that the probe results
reported in the main paper reflect properties of the model's internal
representations rather than artifacts of probe methodology or dataset
construction.

\section{Appendix E. Behavioral Dependence Analyses}

The main paper uses behavioral interventions to complement the representational
probe analyses. This appendix reports the complete behavioral analyses supporting
the conclusion that reasoning content has a measurable but modest causal
influence on intervention decisions after controlling for reasoning-format
effects.

\subsection{E.1 Experimental Design}
Behavioral dependence is evaluated by comparing intervention decisions under
three conditions: \emph{full reasoning}, in which the complete chain of thought
is available; \emph{partial reasoning}, in which the reasoning is truncated
halfway through generation; and \emph{neutral filler}, in which the removed
reasoning is replaced with length-matched neutral text. The neutral-filler
condition controls for disruption of the reasoning format itself. Comparing
truncation directly with the full chain of thought cannot distinguish between
dependence on reasoning content and dependence on the presence of a reasoning
span.

\subsection{E.2 Raw Flip Rates}
Table~\ref{tab:e-flip} reports the proportion of decisions that differ from the
full-reasoning policy under each intervention.

\begin{table}[h]
\centering
\small
\begin{tabular}{@{}lc@{}}
\toprule
Intervention & Flip rate \\
\midrule
Partial reasoning & 0.277 \\
Neutral filler & 0.403 \\
\bottomrule
\end{tabular}
\caption{Decision flip rates under behavioral interventions.}
\label{tab:e-flip}
\end{table}

Replacing the reasoning with neutral filler produces more decision changes than
truncating the reasoning itself. This indicates that much of the raw flip rate
reflects disruption of the reasoning format rather than removal of reasoning
content.

\subsection{E.3 Content Dependence}
The contribution of reasoning content is estimated by comparing the
partial-reasoning and neutral-filler conditions. Retaining the first half of the
original reasoning consistently pulls decisions back toward the full-reasoning
policy relative to replacing the entire span with unrelated filler. Because the
truncation and filler conditions are evaluated on the same examples, all
statistical comparisons use exact McNemar tests.

\begin{table}[h]
\centering
\small
\begin{tabular}{@{}lcc@{}}
\toprule
Policy & Dependence gap & McNemar $p$ \\
\midrule
Base policy & $+0.126$ & $7.0\times10^{-8}$ \\
RL accuracy (seed 1) & $+0.046$ & $0.020$ \\
RL accuracy (seed 2) & $-0.070$ & $0.0015$ \\
RL accuracy (seed 3) & $+0.110$ & $2.5\times10^{-7}$ \\
RL faithfulness (seed 1) & $+0.080$ & $1.2\times10^{-4}$ \\
RL faithfulness (seed 2) & $+0.110$ & $8.7\times10^{-6}$ \\
RL faithfulness (seed 3) & $+0.063$ & $0.0013$ \\
\bottomrule
\end{tabular}
\caption{Filler-controlled behavioral dependence.}
\label{tab:e-dep}
\end{table}

For the base reasoning policy, the dependence gap is consistently positive,
indicating that reasoning content contributes to the final intervention decision
beyond the reasoning format alone.

\subsection{E.4 Decision Stratification}
To determine whether behavioral dependence differs between intervention decisions
and abstentions, we stratify the analysis by the model's original prediction.
Nearly all observed decision changes move toward intervention, producing an
asymmetric distribution of flip opportunities. Consequently, items originally
classified as speak have relatively little opportunity to change under
perturbation, whereas silent decisions can change in either direction. The
resulting asymmetry makes raw dependence substantially larger for silent
decisions without implying that intervention decisions are unsupported by
reasoning. This analysis indicates that behavioral dependence should be
interpreted jointly with the direction of decision changes rather than with flip
rates alone.

\subsection{E.5 Across-Seed Variability}
The RL experiments were repeated using three independent random seeds for each
reward objective. Although each individual run exhibits statistically significant
behavioral dependence, the magnitude of the dependence varies substantially
across training runs. In particular, the accuracy-only reward produces both
positive and negative dependence gaps across different seeds. The faithfulness
reward shows a more consistent sign across runs, although the overall magnitude
remains similar to the base policy. These observations motivated treating the
training run, rather than the individual evaluation example, as the experimental
unit when comparing reinforcement-learning objectives.

\subsection{E.6 Summary}
The behavioral analyses support three conclusions. First, raw chain-of-thought
flip rates substantially overestimate reasoning dependence because disrupting the
reasoning format alone changes model behavior. Second, comparing partial
reasoning with neutral filler isolates the contribution of reasoning content and
reveals a modest but statistically reliable dependence for the base reasoning
policy. Finally, reinforcement-learning objectives produce substantial
run-to-run variability, indicating that objective-level conclusions require
replication across independent training runs rather than inference from a single
optimization trajectory.

\section{Appendix F. Stated Versus Internal Reasons}

The main paper compares the reasons stated in the model's generated chain of
thought with the reasons represented internally by the model. This appendix
provides additional methodological details and complete analyses supporting that
comparison.

\subsection{F.1 Experimental Design}
A central question in faithfulness research is whether the explanation a model
produces reflects the computation that actually influenced its decision
\citep{jacovi2020}. We investigate this question by comparing two representations
of intervention type: the gold intervention type associated with the
conversation, recovered from the source dataset; and the stated intervention type
extracted from the model's generated reasoning. Agreement between these
representations is evaluated using linear probes trained on hidden activations at
different positions in the reasoning process. Unlike the decision probes
presented in the main paper, this analysis focuses on \emph{why} the model claims
an intervention is necessary rather than \emph{whether} it chooses to intervene.

\subsection{F.2 Extracting Stated Reasons}
The stated intervention type is extracted automatically from each generated
reasoning trace using GPT-5 as a classifier. The classifier assigns one of the
five intervention categories used throughout the paper: factual correction,
concept definition, data provision, source identification, and synthesis or
reframing.

We validate this annotation against two human annotators who independently
labelled all 120 sampled traces using the same five categories plus a ``no clear
type'' option, blind to the classifier's output and to each other. Each agrees with the
classifier at a level typical of a five-way scheme (annotator~1 Cohen's
$\kappa = 0.42$, 95\% CI $[0.31, 0.53]$, 55.0\% raw agreement; annotator~2
$\kappa = 0.53$, $[0.41, 0.64]$, 65.8\% raw agreement; $n = 120$ throughout).
The classifier never assigns the ``none'' category, which annotator~1 used on 7
of 120 items and annotator~2 on 1.

The two annotators, however, agree with each other substantially less than either agrees with the classifier: $\kappa = 0.21$, 95\% CI $[0.10, 0.32]$, with
38.3\% raw agreement. Excluding items either marked ``none'' barely changes this
($\kappa = 0.23$, $n = 112$). All three sources select the same category on 41 of
120 items, exactly two agree on 68, and all three differ on 11. The five-way
intervention-type taxonomy is therefore not reliably applicable by humans, and
the residual human-classifier disagreement reflects genuine boundary ambiguity
rather than classifier error. Stated-type results should be read accordingly:
absolute agreement values are unreliable, and because the same annotation
procedure is applied to every policy, only the relative comparison across
policies is interpretable, with small differences falling within annotation noise.

\subsection{F.2.1 Annotator Agreement Analysis}
Table~\ref{tab:f-kappa} reports the label distributions underlying the agreement
statistics. Both annotators additionally had access to a ``none'' option, which
the classifier cannot produce.

\begin{table}[h]
\centering
\small
\begin{tabular}{@{}lccc@{}}
\toprule
Intervention type & Annot.\ 1 & Annot.\ 2 & Classifier \\
\midrule
Data provision & 39 & 45 & 46 \\
Factual correction & 27 & 22 & 21 \\
Concept definition & 18 & 10 & 10 \\
Synthesis \& reframing & 16 & 36 & 34 \\
Source identification & 13 & 6 & 9 \\
None & 7 & 1 & 0 \\
\midrule
Total & 120 & 120 & 120 \\
\bottomrule
\end{tabular}
\caption{Label distributions for the two human annotators and the automatic
classifier on the 120-item validation sample. Annotator~2 applies \emph{synthesis
or reframing} at close to the classifier's rate (36 vs.\ 34) where annotator~1
uses it half as often (16), which accounts for most of the difference between
their agreement scores. The largest annotator--annotator disagreements are data
provision labelled as synthesis or reframing (13 items), and concept definition
and factual correction each labelled as data provision (9 items apiece).}
\label{tab:f-kappa}
\end{table}

Excluding items marked ``none'' changes little: annotator~1's agreement with the
classifier rises only to $\kappa = 0.46$, and annotator--annotator agreement to
$\kappa = 0.23$. The disagreement is therefore not attributable to a single
missing category. We report this as a limitation of the taxonomy as operationalized rather than of the classifier: an automatic labeller cannot exceed the reliability of the scheme it is asked to apply.

\subsection{F.3 Agreement Between Gold and Stated Reasons}
The proportion of examples for which the generated explanation matches the gold
intervention type increases modestly after reinforcement learning.

\begin{table}[h]
\centering
\small
\begin{tabular}{@{}lc@{}}
\toprule
Policy & Agreement \\
\midrule
Base policy & 0.378 \\
RL (accuracy) & 0.425 \\
RL (faithfulness) & 0.474 \\
\bottomrule
\end{tabular}
\caption{Agreement between stated and gold intervention types.}
\label{tab:f-agree}
\end{table}

Although reinforcement learning improves agreement between generated explanations
and the gold labels, agreement alone does not establish that the generated
explanation reflects the computation used to produce the decision.

\subsection{F.4 Probe Position Determines Interpretation}
To distinguish internally represented reasons from generated explanations, we
probe hidden activations at two locations. \emph{Pre-CoT}: activations
immediately before reasoning begins. \emph{End-of-CoT}: activations after the
reasoning has been generated. The two positions support different
interpretations. Before reasoning, any correspondence between probe outputs and
stated explanations must arise from internal model representations because the
explanation has not yet been generated. After reasoning, the residual stream
contains the generated explanation itself. High agreement at this position may
therefore reflect information copied from the generated text rather than an
internal reason that existed before generation.

\subsection{F.5 Pre-CoT Versus End-of-CoT Alignment}
At the end of the chain of thought, probe predictions align significantly more
closely with the stated intervention type than with the gold intervention type.
However, this alignment largely disappears before reasoning begins.

\begin{table}[h]
\centering
\small
\setlength{\tabcolsep}{4pt}
\begin{tabular}{@{}llccc@{}}
\toprule
Policy & Position & Gold & Stated & Perm.\ $p$ \\
\midrule
Base & Pre-CoT & 0.348 & 0.258 & 0.311 \\
Base & End-of-CoT & 0.202 & 0.404 & 0.001 \\
RL (acc.) & Pre-CoT & 0.202 & 0.512 & 0.072 \\
RL (acc.) & End-of-CoT & 0.143 & 0.512 & 0.102 \\
RL (faith.) & Pre-CoT & 0.366 & 0.293 & 0.288 \\
RL (faith.) & End-of-CoT & 0.195 & 0.390 & 0.006 \\
\bottomrule
\end{tabular}
\caption{Alignment between probe predictions and intervention types on the
disagreement subset (all policies). Each cell is the fraction of disagreement
cases whose probe prediction matches the given type; $p$ is the permutation test
for the stated-versus-gold gap.}
\label{tab:f-align}
\end{table}

This pattern suggests that the stated explanation is constructed during reasoning
rather than decoded from an already formed internal representation.

\subsection{F.6 Disagreement Analysis}
The most informative examples are those in which the generated explanation
disagrees with the gold intervention type. Approximately 85 examples satisfy this
criterion for each policy (base 89, accuracy 84, faithfulness 82). To analyze
these disagreements, we train probes only on examples where stated and gold
intervention types agree and evaluate them on the disagreement subset. Because
this subset is relatively small, probe confidence intervals are correspondingly
wide. Across all policies, probe reliability on the disagreement subset remains
modest, limiting the strength of conclusions that can be drawn from individual
disagreement examples.

\subsection{F.7 Methodological Implications}
This analysis highlights an important limitation of dual-probe approaches to
reasoning faithfulness. Before reasoning begins, there may be little or no
explicit representation of the eventual explanation to decode. After reasoning
ends, the model has already generated the explanation, allowing probes to recover
information copied directly from the generated text. Consequently, agreement
between end-of-CoT probes and generated explanations should not be interpreted as
evidence that those explanations faithfully reflected the model's internal
reasoning before generation. Instead, probe position must be considered part of
the experimental design.

\subsection{F.8 Summary}
The stated-versus-internal reason analysis supports three conclusions. First,
reinforcement learning modestly increases agreement between generated
explanations and the gold intervention labels. Second, the strongest alignment
between probe outputs and stated explanations occurs after reasoning has already
been generated. Finally, the absence of comparable alignment before reasoning
suggests that stated explanations are constructed during reasoning rather than
read directly from a pre-existing internal representation. These findings
motivate treating probe position as a first-order consideration when interpreting
dual-probe faithfulness analyses.

\section{Appendix G. Single-Run Reward Comparisons Can Be Misleading}

The main paper compares reinforcement learning optimized for decision accuracy
with reinforcement learning that additionally rewards behavioral dependence on
the model's own reasoning. This appendix reports the complete reward-ablation
analyses and examines the variability introduced by stochastic reinforcement
learning.

\subsection{G.1 Reward Objectives}
We compare two reinforcement-learning objectives. The accuracy objective rewards
only the correctness of the final intervention decision; the faithfulness
objective augments this reward with the behavioral dependence bonus described in
Appendix~A.6. Both objectives begin from the same instruction-tuned checkpoint
and use identical optimization settings. The only difference between conditions
is the additional behavioral dependence reward.

\subsection{G.2 Behavioral Evaluation}
Table~\ref{tab:g-behav} reports the complete behavioral evaluation for both
reward objectives.

\begin{table}[h]
\centering
\small
\begin{tabular}{@{}lccc@{}}
\toprule
Policy & Macro-F1 & AUROC & Dependence gap \\
\midrule
Base think & 0.536 & 0.637 & $+0.126$ \\
RL (accuracy) & 0.520 & 0.640 & $+0.046$ \\
RL (faithfulness) & 0.500 & --- & $+0.080$ \\
\bottomrule
\end{tabular}
\caption{Behavioral evaluation of reinforcement-learning objectives. The
faithfulness objective was trained with two dependence counterfactuals, yielding
two distinct checkpoints. Table 1 of the main paper reports the empty-think
variant (0.549, AUROC 0.622); the row above is the truncation-dependence variant
(0.500, AUROC not measured on the enriched slice), which all behavioral and probe
analyses use.}
\label{tab:g-behav}
\end{table}

Across all deployment metrics, both reinforcement-learning objectives remain
statistically similar to the base reasoning policy. The targeted behavioral
quantity also changes only modestly. Although the faithfulness objective produces
a dependence gap with a consistent direction across training runs, its magnitude
remains comparable to that of the base policy.

\subsection{G.3 Complete Seed Results}
Because reinforcement learning uses stochastic sampling throughout optimization, a
single training run may not accurately represent the behavior of a reward
objective. To estimate this variability, we independently repeat both
reinforcement-learning objectives using three random seeds.

\begin{table}[h]
\centering
\small
\begin{tabular}{@{}llc@{}}
\toprule
Objective & Seed & Dependence gap \\
\midrule
Accuracy & 1 & $+0.046$ \\
Accuracy & 2 & $-0.070$ \\
Accuracy & 3 & $+0.110$ \\
Faithfulness & 1 & $+0.080$ \\
Faithfulness & 2 & $+0.110$ \\
Faithfulness & 3 & $+0.063$ \\
\bottomrule
\end{tabular}
\caption{Behavioral dependence across independent training runs.}
\label{tab:g-seeds}
\end{table}

Although every individual run exhibits statistically significant behavioral
dependence relative to its own perturbation controls, the magnitude varies
substantially across optimization runs. For the accuracy objective, the estimated
effect even changes sign across seeds.

\subsection{G.4 Between-Run Variability}
\begin{figure}[h]
\centering
\includegraphics[width=\columnwidth]{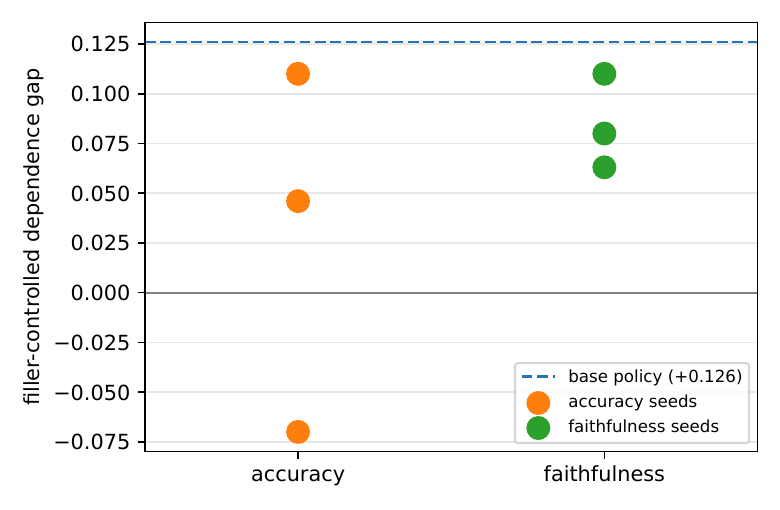}
\caption{Behavioral dependence across independent reinforcement-learning runs.
Between-run variability exceeds the observed differences between reward
objectives.}
\label{fig:g-runs}
\end{figure}

The variability across optimization runs exceeds the differences observed between
reward objectives. Consequently, conclusions drawn from a single
reinforcement-learning run may reflect optimization noise rather than properties
of the reward function itself. Figure~\ref{fig:g-runs} summarizes the observed
distribution of behavioral dependence across all training runs.

\subsection{G.5 Implications for Reward Comparisons}
This result has methodological implications beyond the present task. Many recent
reinforcement-learning studies compare reward objectives using a single
optimization run per condition. In stochastic optimization settings, however,
individual evaluation examples are not independent observations of the reward
objective. Instead, the optimization run itself is the experimental unit.
Item-level statistical tests therefore cannot establish differences between
reward objectives when only a single optimization trajectory is available.
Objective-level comparisons require replication across independent training runs.

\subsection{G.6 Interpretation}
The reward-ablation experiments support two conclusions. First, adding a
behavioral dependence reward does not substantially improve any deployment or
audit metric relative to the base reasoning policy. Second, the variability
introduced by stochastic optimization exceeds the measured differences between
reward objectives. These findings suggest that objective-level claims should be
interpreted cautiously unless they are supported by multiple independent training
runs.

\subsection{G.7 Summary}
Across all evaluated audits, reinforcement learning optimized for behavioral
dependence performs similarly to reinforcement learning optimized only for
decision accuracy. The primary methodological finding is therefore not that one
reward objective outperforms another, but that between-run variability can exceed
the apparent differences between objectives. This motivates treating the
optimization run, rather than the individual evaluation example, as the
appropriate unit of inference when comparing reinforcement-learning objectives.

\end{document}